\documentclass[times, review, 10pt]{elsarticle}

\usepackage{amssymb}
\usepackage{amsmath} % for align, equation, etc.
\usepackage{framed,multirow}
\usepackage{latexsym}
\usepackage{textcomp,booktabs}
\usepackage{tabularx}
\usepackage{makecell}
\usepackage{amssymb}
\usepackage{pifont}
\usepackage{url}
\usepackage{xcolor}
\definecolor{newcolor}{rgb}{.8,.349,.1}
\usepackage{marvosym}
\usepackage{hyperref}
\usepackage{amsmath}
\usepackage{amssymb}
\usepackage{bm}          % For bold math symbols (vectors)
\usepackage{algorithm}   % For algorithm floats
\usepackage{algpseudocode} % For algorithmic environments
\usepackage{cleveref}    % Optional: for clever referencing
\usepackage{float}
\usepackage{graphicx}
\usepackage{booktabs}
\usepackage{multirow}
\usepackage{tabularx}
\usepackage{makecell}
\usepackage{amssymb}
\newcommand{\yes}{\checkmark}
\newcommand{\no}{$\times$}

\journal{Pattern Recognition}

\begin{document}

\begin{frontmatter}

%% Title
\title{UniPINN: A Unified PINN Framework for Multi-task Learning of Diverse Navier-Stokes Equations}

%% Authors and addresses
% \author{Dengdi Sun^{1}}
% \author{Jie Chen^{1}}
% \author{Xiao Wang^{2}\corref{cor1}}
% \author{Jin Tang^{2}}
% \address{1. School of Artificial Intelligence, Anhui University, Hefei 230601, China}
% \address{2. School of Computer Science and Technology, Anhui University, Hefei 230601, China} 

\author[O1]{Dengdi Sun}
\author[O1]{Jie Chen}
\author[O2]{Xiao Wang\corref{mycorrespondingauthor}}
%\texorpdfstring{%}
\cortext[mycorrespondingauthor]{Corresponding author}
\ead{xiaowang@ahu.edu.cn}
\author[O2]{Jin Tang}

%% Author affiliation
\address[O1]{Key Laboratory of Intelligent Computing \& Signal Processing (ICSP), Ministry of Education, School of Artificial Intelligence, Anhui University, Hefei 230601, China}%Department and Organization
\address[O2]{Anhui Provincial Key Laboratory of Multimodal Cognitive Computing, School of Computer Science and Technology, Anhui University, Hefei 230601, China}%Department and Organization

%\texorpdfstring

% \author{Dengdi Sun^{1}, Jie Chen^{1}, Xiao Wang^{2}, Jin Tang^{2} } 
% \address{1. Key Laboratory of Intelligent Computing \& Signal Processing (ICSP), Ministry of Education, School of Artificial Intelligence, Anhui University, Hefei 230601, China}
% \address{2. Anhui Provincial Key Laboratory of Multimodal Cognitive Computing, School of Computer Science and Technology, Anhui University, Hefei 230601, China}

%\cortext[cor1]{Corresponding author.}

%% Abstract
\begin{abstract}
Physics-Informed Neural Networks (PINNs) have shown promise in solving incompressible Navier-Stokes equations, yet existing approaches are predominantly designed for single-flow settings. When extended to multi-flow scenarios, these methods face three key challenges: (1) difficulty in simultaneously capturing both shared physical principles and flow-specific characteristics, (2) susceptibility to inter-task negative transfer that degrades prediction accuracy, and (3) unstable training dynamics caused by disparate loss magnitudes across heterogeneous flow regimes. To address these limitations, we propose UniPINN, a unified multi-flow PINN framework that integrates three complementary components: a shared-specialized architecture that disentangles universal physical laws from flow-specific features, a cross-flow attention mechanism that selectively reinforces relevant patterns while suppressing task-irrelevant interference, and a dynamic weight allocation strategy that adaptively balances loss contributions to stabilize multi-objective optimization. Extensive experiments on three canonical flows demonstrate that UniPINN effectively unifies multi-flow learning, achieving superior prediction accuracy and balanced performance across heterogeneous regimes while successfully mitigating negative transfer. The source code of this paper will be released on \href{https://github.com/Event-AHU/OpenFusion}{https://github.com/Event-AHU/OpenFusion}. 
%In recent years, Physics-Informed Neural Networks (PINNs) have been used for flow simulations of various incompressible Navier-Stokes equations, but single-task PINNs have inherent limitations: low data efficiency under sparse supervision, weak generalization ability across flow patterns, and high computational costs due to repeated training. Furthermore, inter-task negative transfer constrains their performance in complex flow simulations. To tackle these challenges, this study proposes a multi-task PINN framework that combines cross-flow attention with dynamic weight allocation (DWA) strategies. The aim is to verify the efficiency of this framework in various typical flows, focusing on alleviating negative transfer, enhancing training stability across different viscosities, and exploring the range of advantages. Extensive experiments demonstrate the effectiveness of our approach. Especially, with the cross-flow attention module, inter-task negative transfer is significantly mitigated, enabling flow tasks with advantages to maintain stable training curves; the introduction of dynamic weight allocation further reduces loss oscillations during training, notably enhancing the convergence speed of certain flow tasks. 
\end{abstract}

%% Keywords

\end{frontmatter}

\section{Introduction}
Physics-Informed Neural Networks (PINNs) have demonstrated revolutionary potential in solving partial differential equations (PDEs) since their inception~\citep{ref1_raissi2019physics}, particularly achieving remarkable success in the field of fluid mechanics. The core innovation of PINNs lies in the seamless integration of physical laws, such as the conservation of momentum and mass, directly into the neural network's loss functions. By leveraging automatic differentiation to compute high-order derivatives, PINNs ensure that the network satisfies fundamental physical constraints during training. This mesh-free paradigm enables the learning of complex fluid dynamics behaviors from sparse observational data and facilitates the handling of intricate geometries and boundary conditions that have traditionally posed challenges for conventional numerical methods~\citep{fang2024learning}. To date, PINNs have been successfully applied to a variety of canonical fluid problems, including lid-driven cavity flow~\citep{ref2_bai2020applying}, pipe flow~\citep{ref3_urbanowicz2023navier}, and Couette flow~\citep{ref4_mehta2019discovering}. These models have proven capable of accurately capturing vortex structures~\citep{jiang2025gradient}, predicting laminar-to-turbulent transitions~\citep{hanrahan2023predicting}, and resolving high-fidelity velocity distributions and pressure fields across various reynolds numbers~\citep{ren2024physics}, thereby underscoring their tremendous potential in computational fluid dynamics (CFD).

\begin{figure}[t]
    \centering
    \includegraphics[width=1\linewidth]{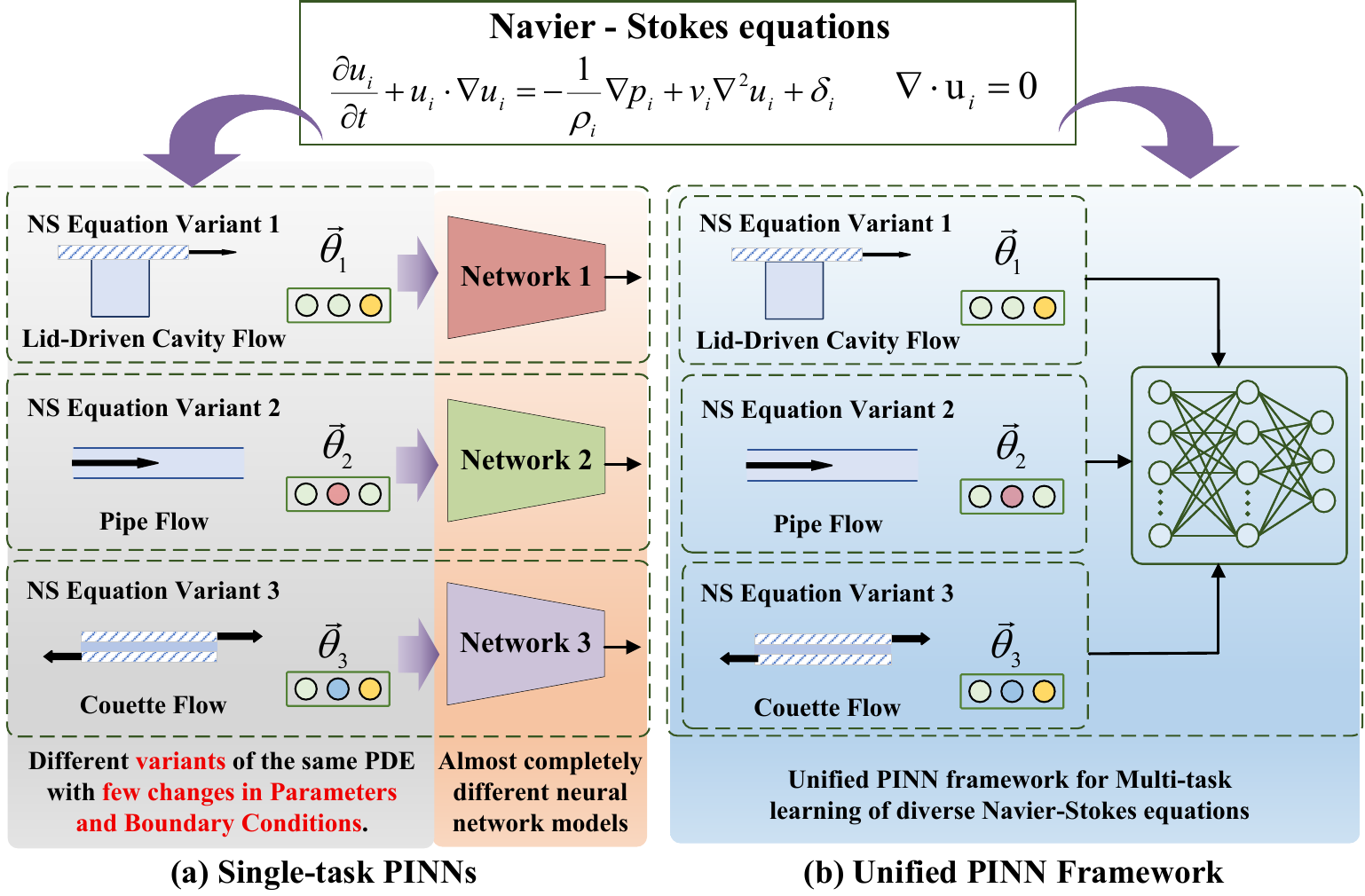}
    \caption{Single-task vs.\ unified multi-flow learning.
    (a) Conventional single-task PINNs train independent networks for each flow type, incurring parameter redundancy and neglecting the shared Navier-Stokes structure across flows.
    (b) UniPINN unifies multiple flow types under a shared backbone with task-specific heads and cross-flow attention, enabling joint learning and knowledge transfer while preserving flow-specific fidelity.}
    \label{fig:unipinn_framework}
\end{figure}

However, real-world fluid problems typically involve a diverse array of flow regimes, each characterized by unique physical properties and boundary conditions, which pose significant challenges for the standardized application of PINNs. In such settings, multi-flow learning must contend with substantial discrepancies in physical parameters, flow field characteristics, and geometric configurations. Variations in parameters such as viscosity, density, and characteristic velocity fundamentally alter the relative dominance of convective and diffusive terms within the Navier-Stokes equations, leading to divergent flow behaviors~\citep{ref5_malek2005mathematical}. Existing methods often treat these distinct flow types as isolated problems, necessitating the training of independent networks for each scenario~\citep{stiasny2023physics}. This decoupled approach not only incurs prohibitive computational overhead but also neglects the inherent physical similarities and the potential for knowledge transfer between related flow regimes. While disparate flows often share congruent features, such as velocity gradients in boundary layers or pressure distributions at vortex cores~\citep{ref8tanarro2020effect}, these commonalities remain largely unexploited by independently trained architectures. As illustrated in Fig.~\ref{fig:unipinn_framework}, nearly all flow types are governed by the same underlying Navier-Stokes equations, differing primarily in their boundary and initial conditions~\citep{bonfanti2024generalization}. This shared mathematical foundation provides a critical basis for cross-flow knowledge sharing~\citep{zhu2024online}, yet the lack of a unified multi-flow learning framework has hindered the effective utilization of these universal physical principles.

In light of this challenge, a natural starting point would be to draw upon existing multi-task learning frameworks developed in other domains. However, despite the rapid progress of multi-task learning in fields such as computer vision and natural language processing~\citep{ref15zhang2021survey}, these generic architectures are not directly transferable to PINN-based multi-flow modeling. Conventional multi-task methods are typically designed for data-driven objectives where loss scales are often comparable across tasks, and they do not explicitly account for PDE residuals, physical constraints, or the varying numerical stiffness inherent to different flow regimes. Naïvely combining heterogeneous fluid tasks within such frameworks frequently leads to incompatible gradient magnitudes, unstable optimization dynamics, and the degradation of delicate physical structures within individual flows ~\citep{ref14chen2024multi}.

%One might naturally consider leveraging existing multi-task learning frameworks from other domains to address this challenge. However, despite the rapid progress of multi-task learning in computer vision and natural language processing~\citep{ref13ruder2017overview,ref15zhang2021survey}, these generic frameworks cannot be directly transplanted to PINN-based multi-flow modeling. Conventional multi-task methods are typically designed for data-driven objectives with comparable loss scales and do not explicitly account for PDE residuals, physical constraints, or the stiffness of different flow regimes. Naively combining heterogeneous fluid tasks within such architectures often results in incompatible gradient magnitudes, unstable optimization, and the erosion of delicate physical structures in individual flows~\citep{ref12caruana1997multitask,ref14chen2024multi}. 

Moreover, many existing schemes adopt rigid weight-sharing mechanisms, which struggle to simultaneously capture universal physical patterns while preserving flow-specific nuances like boundary layer behavior or vortex shedding characteristics. This structural rigidity hampers generalization when models are required to accommodate heterogeneous flows spanning vastly different physical regimes. At the optimization level, the simultaneous training of diverse flow types within a single network is further complicated by the disparate magnitudes of their respective loss components, which often span several orders of magnitude. Such imbalances lead to conflicting optimization trajectories ~\citep{li2024physics}, a phenomenon commonly referred to as gradient pathology ~\citep{yu2022gradient}. This occurs when the dominant gradient direction associated with one flow regime suppresses the optimization of others, causing the model to satisfy certain physical constraints at the expense of violating equally fundamental ones. As a result, delicate physical structures embedded in individual flows may be eroded or lost entirely during training. Consequently, these limitations underscore the pressing need for a physics-informed multi-task architecture specifically designed for PINN-based multi-flow modeling, not only respects the underlying Navier–Stokes structure across diverse flow types but also actively mitigates gradient pathology to ensure balanced and physically consistent learning.

%many existing multi-task schemes adopt rigid weight-sharing, making it difficult to simultaneously capture universal physical patterns and preserve flow-specific nuances. This structural rigidity limits generalization when dealing with heterogeneous flows across broad physical scales. At the optimization level, the simultaneous training of diverse flow regimes within a single network is further plagued by disparate loss magnitudes, often spanning several orders of magnitude, which lead to conflicting optimization trajectories~\citep{li2024physics}. This phenomenon, often referred to as gradient pathology~\citep{wang2021understanding,yu2022gradient}, arises when the dominant gradient direction of one loss term suppresses the optimization of other physical constraints, causing the model to satisfy some equations while violating others. Consequently, there is a clear need for a physics-aware multi-task architecture that both respects the Navier--Stokes structure and mitigates this gradient pathology when learning across multiple flow types.

To address these problems, this paper proposes \textbf{UniPINN}: a \textbf{Uni}fied multi-task \textbf{P}hysics-\textbf{I}nformed \textbf{N}eural \textbf{N}etwork framework specifically designed to overcome the limitations of conventional multi-task learning in physics-constrained settings. Unlike generic multi-task architectures that struggle with heterogeneous physical regimes, UniPINN introduces a shared-specialized architecture that explicitly balances the trade-off between capturing universal physical principles (e.g., conservation laws embedded in the Navier–Stokes equations) and preserving flow-specific features (e.g., boundary layer dynamics or vortex structures). Complemented by a cross-flow attention mechanism, the framework enables adaptive knowledge sharing across different flow regimes by selectively attending to physically relevant features, thereby mitigating the rigid weight-sharing problem inherent in traditional approaches. To address the optimization challenges posed by disparate loss magnitudes and gradient pathologies, UniPINN further incorporates an adaptive weight balancing strategy that dynamically adjusts the contribution of each flow's physical constraints during training. This ensures that no single flow regime dominates the optimization trajectory, preserving the fidelity of delicate physical structures across all tasks. By embedding the physical constraints of diverse flow regimes into a unified objective function while actively managing parameter discrepancies and gradient conflicts, UniPINN achieves efficient knowledge transfer and robust multi-flow learning. The primary contributions of this work are as follows:

%To address these challenges, this paper proposes UniPINNs, a unified multi-flow Physics-Informed Neural Network framework. By utilizing a shared-specialized architecture and a cross-flow attention mechanism, UniPINNs embeds the physical constraints of diverse flow regimes into a unified objective function. Combined with an adaptive weight balancing strategy, the framework effectively manages parameter discrepancies across flows, achieving efficient knowledge sharing and robust learning. The primary contributions of this work are as follows:

\begin{itemize}
\item We construct a unified learning framework using a shared-specialized architecture that extracts universal physical operator features through shared layers while capturing task-specific boundary nuances through task-specific dedicated heads, significantly enhancing multi-task pre-training efficiency.
\item We design a cross-flow attention mechanism incorporating self-attention and cross-attention modules to facilitate physical knowledge interaction. This mechanism dynamically identifies similar topological structures, such as vortex evolution patterns, effectively mitigating the risks of negative transfer common in traditional methodologies.
\item To resolve gradient conflicts during joint training, we propose an adaptive weight balancing strategy that real-time monitors the training state and residual distribution of each flow regime, dynamically adjusting loss weights to ensure convergence stability across large-scale heterogeneous tasks.
\end{itemize}

\section{Related Works}

\subsection{Physics-Informed Neural Networks}
Since their introduction, Physics-Informed Neural Networks (PINNs) have established a new paradigm for solving forward and inverse problems in fluid dynamics by embedding physical laws directly as regularization terms within the loss function~\citep{ref1_raissi2019physics}. %Since the introduction of PINNs, the paradigm of embedding physical laws as regularization terms has revolutionized the solution of forward and inverse problems in fluid dynamics~\citep{ref1_raissi2019physics}. 
By incorporating PDE residuals directly into the loss function and leveraging automatic differentiation to compute high-order derivatives, PINNs can learn complex fluid behaviors from sparse observational data while respecting mass and momentum conservation~\citep{fang2024learning,MAO2026111816}. Early studies demonstrated that PINNs could effectively solve canonical incompressible Navier-Stokes problems, including lid-driven cavity flow, Poiseuille pipe flow, and Couette flow~\citep{ref3_urbanowicz2023navier}, and handle complex geometries that pose challenges for traditional numerical solvers, motivating a large body of follow-up work on fluid applications~\citep{YU2026112123}.

As the scope of PINN-based modeling has expanded from simple laminar flows to high-Reynolds-number and multiscale regimes, however, the limitations of standard fully-connected architectures have become increasingly apparent. A critical bottleneck is the phenomenon of spectral bias, wherein neural networks prioritize learning low-frequency functions at the expense of high-frequency physical features such as turbulent microstructures or shock waves~\citep{zhao2024comprehensive}. This limitation has motivated two complementary lines of research. The first line explores extensions to more complex fluid problems, including multi-regime and transient scenarios. The second line focuses on architectural innovations to overcome structural deficiencies in the underlying networks, such as deeper and more adaptive designs. For instance, \citep{wang2024piratenets} introduced PirateNets with learnable adaptive weights in skip connections to alleviate vanishing gradients, while other variants have explored adaptive activation functions, modified depth configurations, and enhanced residual structures to better capture challenging physical regimes~\citep{ref27duibiLAAFxu2025physics}.

Despite these advances, a fundamental limitation persists across both research directions: existing approaches predominantly treat each flow type independently~\citep{ref6_hennigh2021nvidia}. Whether through problem-specific extensions or architectural improvements, current methods remain inherently single-task in nature, failing to exploit the shared physical structure that underlies diverse flow regimes. While disparate flows often share common features, such as boundary layer dynamics, vortex structures, and the conservation laws embedded in the Navier–Stokes equations, these commonalities remain largely unexploited due to the lack of unified frameworks capable of learning across heterogeneous flow regimes. What is needed is an approach that enables knowledge transfer between related flows, leverages cross-flow physical commonalities, and simultaneously preserves the distinct characteristics and boundary conditions unique to each flow type. The shared-specialized architecture proposed in this work addresses precisely this gap, enabling joint multi-flow learning through a common backbone augmented with flow-specific heads and a cross-flow attention mechanism that supports knowledge transfer while maintaining flow-specific fidelity.

\subsection{Multi-Task Learning and Gradient Balancing}
Applying PINNs to complex fluid systems, such as fluid-structure interaction or multi-regime networks~\citep{moya2024multi}, inherently constitutes a multi-objective optimization problem. In such scenarios, the residuals of the Navier-Stokes equations, boundary condition residuals, and initial condition residuals often differ by several orders of magnitude. This imbalance gives rise to severe gradient pathology~\citep{wang2021understanding}, where the dominant gradient direction of one loss term suppresses the optimization of other physical constraints, resulting in a model that satisfies some equations while violating others.

To reconcile these conflicts, Multi-Task Learning (MTL) approaches have been increasingly adopted. Classical MTL studies have shown that sharing representations across related tasks can improve generalization and data efficiency~\citep{LI2026108694}. Recent surveys and empirical studies further systematize these ideas, highlighting both the benefits and the risks of negative transfer when tasks are insufficiently related or poorly balanced~\citep{JI2025111423}. In response, a variety of uncertainty-weighted and adaptive multi-task strategies have been proposed to automatically adjust task contributions, often by modeling task uncertainty or gradient statistics~\citep{ref16chen2025multi}.

Within scientific computing and PINN-based modeling, these ideas above have been extended to handle coupled physical fields and multiple objectives. For example, \citep{li2024multitask} used soft parameter sharing for fluid-structure interaction, and \citep{zhang2025self} proposed a self-adaptive weighting algorithm to rebalance losses during training. Other works have focused on controlling gradient magnitudes or aligning gradient directions across tasks,
% Other works control gradient magnitudes or align gradient directions across tasks
with uncertainty-based and gradient-normalization strategies also gaining traction. Despite this progress, most of these methods are designed for supervised learning settings or particular coupled systems~\citep{subramanian2024adaptive} and rarely integrate adaptive balancing with physics-informed constraints and shared representations across multiple heterogeneous flows. This motivates the need for unified schemes that coordinate gradients at both the task level and the physical law level. %leaving room for unified schemes that coordinate gradients at the level of both tasks and physical laws. 
The dynamic weight allocation (DWA) strategy proposed in this work addresses precisely this challenge. By adaptively reweighting loss terms in real time, DWA ensures that no single objective dominates the optimization trajectory while maintaining alignment with the shared backbone and cross-flow attention mechanism. This enables coordinated gradient updates across multiple flows, preserving the fidelity of delicate physical structures in each flow while still benefiting from cross-flow knowledge transfer.
%The dynamic weight allocation (DWA) strategy proposed in this work solves the problem of gradient imbalance when coordinating both task-level and physics-level objectives across multiple flows, by adaptively reweighting loss terms in real time so that no single objective dominates while maintaining alignment with the shared backbone and cross-flow attention.

\subsection{Physics-Aware Attention Mechanisms}
Traditional Convolutional Neural Networks (CNNs) and fully connected networks typically possess spatial translation invariance or global receptive fields, which limits their specificity when handling local singularities in fluids, such as boundary layer separation or corner vortices~\citep{ref8tanarro2020effect}. In contrast, attention mechanisms introduced in the Transformer architecture and later adapted to vision tasks~\citep{ref21mauricio2023comparing} have proven effective at capturing long-range dependencies and focusing on salient regions, and have recently been brought into scientific machine learning to enhance physical representation capabilities~\citep{peng2023linear}.

Unlike applications in natural language processing and vision, physics-driven attention mechanisms focus on identifying high-error or physically critical regions within the spatiotemporal domain. \citep{aizpurua2025residual} used PDE residual distributions as priors to guide attention toward high-error regions. \citep{li2024transformer} applied cross-attention in Transformer-based physics-informed networks for zero-shot generalization across varying physical conditions. These efforts illustrate how attention can be tailored to physical structure and constraints~\citep{hao2023gnot}, and they provide the theoretical basis for physics-aware attention in scientific machine learning. Nonetheless, most existing approaches are still confined to single-modality or single-task scenarios, focusing on either local error refinement or single-flow mappings in isolation~\citep{ref10zhao2024comprehensive}. There remains a need for attention mechanisms that explicitly support cross-flow knowledge sharing while suppressing negative transfer within unified physics-informed architectures that handle multiple heterogeneous flows simultaneously~\citep{YANG2026111868}. The cross-flow attention module proposed in this work solves the problem of lacking attention mechanisms that support cross-flow knowledge sharing while suppressing negative transfer by enabling selective aggregation of relevant patterns from other flows and filtering out task-incompatible features within a unified physics-informed architecture.

\section{Methodology}
\subsection{Problem Definition}
We consider the problem of learning multiple incompressible Navier-Stokes flow patterns simultaneously using Physics-Informed Neural Networks. Given a set of flow types $\mathcal{F} = \{f_1, f_2, \ldots, f_N\}$, where $N$ is the total number of flow types and the index $i \in \{1, \ldots, N\}$ denotes a generic flow type, each $f_i$ is characterized by its specific physical parameters (e.g., viscosity coefficient $v_i$, density $\rho_i$) and boundary conditions. Our goal is to train a unified neural network that can accurately predict the velocity field $\mathbf{u}_i(x,t)$ and pressure field $p_i(x,t)$ for any flow type $f_i$. In the following, each flow type is treated as one task, and we use these terms interchangeably.

For each flow type $f_i$, the governing equations are the incompressible Navier-Stokes equations~\citep{ref29gu2024physics}:
\begin{align}
\frac{\partial \mathbf{u}_i}{\partial t} + \mathbf{u}_i \cdot \nabla \mathbf{u}_i &= -\frac{1}{\rho_i}\nabla p_i + v_i \nabla^2 \mathbf{u}_i + \bm{\delta}_i, \label{eq:ns_momentum} \\
\nabla \cdot \mathbf{u}_i &= 0, \label{eq:ns_continuity}
\end{align}
where $\mathbf{u}_i(x,t) \in \mathbb{R}^2$ is the velocity field of flow type $f_i$ at spatial position $x = (x,y)$ and time $t$, $p_i(x,t) \in \mathbb{R}$ is the pressure field, $\bm{\delta}_i(x,t) \in \mathbb{R}^2$ is the body force, $\rho_i \in \mathbb{R}^+$ is the density, $v_i \in \mathbb{R}^+$ is the kinematic viscosity coefficient, and the spatial domain $\Omega_i \subset \mathbb{R}^2$ and time domain $[0,T_i] \subset \mathbb{R}^+$ are flow-type specific.

Equation (\ref{eq:ns_momentum}) represents the momentum conservation law, containing local acceleration, convective acceleration, pressure gradient force, viscous force, and external body force. Equation (\ref{eq:ns_continuity}) represents the mass conservation law, i.e., the incompressibility condition. Although all flow types share the same underlying physical laws encoded in \eqref{eq:ns_momentum}-\eqref{eq:ns_continuity}, they exhibit markedly different behaviors due to variations in physical parameters, boundary conditions, and geometric configurations. This diversity poses fundamental challenges for multi-flow learning that go beyond conventional single-task PINNs. Specifically, designing a unified framework that can simultaneously accommodate multiple flow regimes requires addressing the following interrelated challenges:
\begin{itemize}
\item \textbf{Preserving flow-specific features while capturing shared physics:} The model must disentangle universal physical principles (e.g., conservation laws) from flow-specific characteristics (e.g., boundary layer profiles, vortex dynamics) to enable effective knowledge transfer without sacrificing fidelity to individual flows.
\item \textbf{Handling heterogeneous physical parameters and boundary conditions:} Different flow types may operate under vastly different Reynolds numbers, domain geometries, and forcing terms. The framework must flexibly adapt to such variations without requiring separate networks.
\item \textbf{Balancing multiple loss terms across diverse flow types:} The residuals of the Navier-Stokes equations, boundary conditions, and initial conditions for each flow type often exhibit large magnitude discrepancies. These imbalances can lead to gradient pathology, where optimization becomes dominated by a subset of tasks at the expense of others.
\item \textbf{Enabling knowledge transfer across flow types:} While related flows share common fluid mechanical features, existing single-task PINNs fail to exploit these commonalities. A unified approach must facilitate positive transfer: leveraging shared representations to improve data efficiency and generalization, while avoiding negative interference.
\end{itemize}
These challenges motivate the architecture and training strategy proposed in the following subsections.

%The key challenges lie in designing a unified framework that can:
%(1) Capture flow-type specific features while sharing common physical principles,
%(2) Effectively handle different physical parameters and boundary conditions,
%(3) Balance multiple loss terms across different flow types,
%(4) Enable knowledge transfer between flow types.

\begin{figure}[H]
    \centering
    \includegraphics[width=1.0\linewidth]{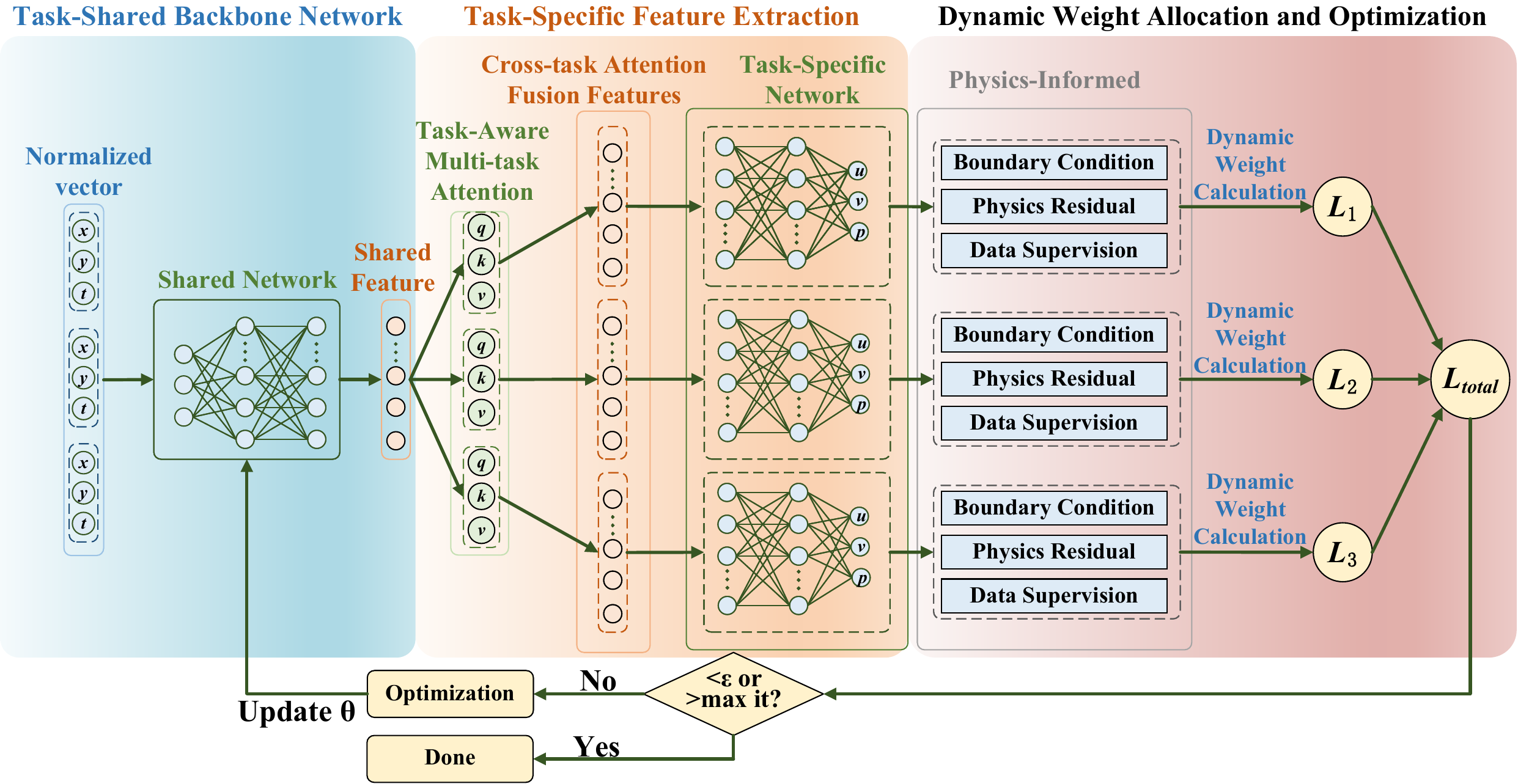}
    \caption{UniPINN architecture. It consists of three components:
    (1) Task-shared backbone network: extracts common physical features from spatiotemporal coordinates and task embeddings for all flow types.
    (2) Task-specific feature extraction: the task-aware multi-task attention selectively reinforces the shared representation; the task-specific dedicated layers decode the enhanced features into stream function and pressure, with velocity obtained via automatic differentiation.
    (3) Dynamic weight allocation: adaptively reweights equation, boundary, and data loss terms across flows to ensure stable joint optimization.}
    \label{fig:unipinn_arch}
\end{figure}

\subsection{Unified Multi-Flow Learning Framework}

Building upon the problem formulation in the previous subsection, we now introduce the overall UniPINN framework, as illustrated in Fig.~\ref{fig:unipinn_arch}. This framework adopts a shared--specialized architecture that couples a task-shared backbone network with task-specific attention and output heads, enabling heterogeneous flows to be modeled within a single unified architecture. At an initial stage, shared layers extract universal physical features from the input coordinates and task embeddings, while the subsequent attention modules and dedicated heads perform task-aware feature refinement and decoding. The remainder of this subsection details each component in turn, including the task-shared backbone, task-specific feature extraction modules, and the dynamic weight allocation strategy for multi-flow optimization.

\subsubsection{Task-Shared Backbone Network}

The shared backbone network $\mathcal{N}_s$ is the core of the entire framework, responsible for learning common physical features across all flow types. This network adopts a Multilayer Perceptron (MLP) structure. For each flow type $f_i$, we introduce a learnable embedding vector $\mathbf{e}_i \in \mathbb{R}^{d_e}$ to encode the latent attributes of the flow type during training, where $d_e$ is the embedding dimension. The spatiotemporal coordinates $(x, y, t)$ and task embeddings are concatenated and fed into the shared MLP:
\begin{align}
\mathbf{x}_{in} = [x, y, t, \mathbf{e}_i] \in \mathbb{R}^{3 + d_e},
\end{align}
The shared backbone network consists of $L$ hidden layers with smooth nonlinear activation functions $\sigma$ (such as $\tanh$ or Swish) to guarantee the existence of high-order derivatives required for calculating PDE residuals:
\begin{align}
& \mathbf{h}^{(0)} = \mathbf{x}_{in}, \\
& \mathbf{h}^{(l)} = \sigma(\mathbf{W}^{(l)} \mathbf{h}^{(l-1)} + \mathbf{b}^{(l)}), \quad l = 1, 2, \ldots, L, \\
&\mathbf{h} = \mathbf{h}^{(L)} = \mathcal{N}_s(\mathbf{x}_{in}; \Theta_{shared}) \in \mathbb{R}^{d_h},
\end{align}
where $\mathbf{W}^{(l)} \in \mathbb{R}^{d_l \times d_{l-1}}$ and $\mathbf{b}^{(l)} \in \mathbb{R}^{d_l}$ are the weight matrix and bias vector of the $l$-th layer, $\Theta_{shared} = \{\mathbf{W}^{(l)}, \mathbf{b}^{(l)}\}_{l=1}^{L}$ denotes the shared network parameters, and $d_h$ is the feature dimension. The output feature $\mathbf{h}$ is a shared representation containing multiple possible physical modes within the shared feature space, capturing universal fluid dynamics patterns such as mass conservation, momentum conservation, and other fundamental physical principles. The design of the shared network enables different flow types to share these common features, improving learning efficiency and facilitating knowledge transfer across heterogeneous fluid tasks.

\subsubsection{Task-Specific Feature Extraction}
Since the shared feature $\mathbf{h}$ inevitably encodes multiple physical modes arising from diverse flow types, directly feeding this entangled representation into separate prediction heads for each flow introduces interference from irrelevant patterns. Such interference can lead to negative transfer, where knowledge from unrelated flows degrades rather than improves prediction performance.
%Since the shared feature $\mathbf{h}$ may entangle multiple potential physical modes, directly using it for a specific flow task can introduce interference from irrelevant patterns, and not all modes are informative for the current flow regime.
To address this challenge, we design a task-specific feature extraction module that maps shared features to flow-specific predictions through the cooperation of a task-aware multi-task attention mechanism and task-specific dedicated layers. Specifically, first, the task-aware multi-task attention mechanism establishes differentiable interactions between the shared representation and each target flow, selectively reinforcing physically relevant patterns while suppressing information that may be detrimental to the current flow regime. Second, each flow type $f_i$ possesses the independent dedicated layer $\mathcal{N}_i$, which decodes this enhanced representation into the corresponding physical field predictions. By combining selective attention with flow-specific decoding, this design preserves the fidelity of individual flow characteristics while still benefiting from shared knowledge across regimes.

%establishes a differentiable interaction and transfer pathway across flow types, selectively reinforcing and reorganizing the shared representation; the dedicated layers then decode the enhanced features into the corresponding physical field predictions. This design aims to mitigate negative transfer. 

\vspace{6pt}
\noindent\textbf{1) Task-aware Multi-task Attention Mechanism} 
\vspace{1pt}

The task-aware multi-task attention mechanism serves as a differentiable feature selection module positioned between the shared backbone and the task-specific layers. This mechanism achieves knowledge interaction and transfer between flow types through carefully designed self-attention and cross-attention modules, effectively capturing common features and differences between flow types. To facilitate learning physical features at different scales and patterns, we adopt a multi-head parallel processing strategy.

\vspace{1pt}
\textbf{\ding{172} Self-Attention Module.}
For each flow type $f_i$, the attention mechanism generates query, key, and value vectors $(\mathbf{Q}_i, \mathbf{K}_i, \mathbf{V}_i)$ from the shared feature $\mathbf{h}$. The self-attention module first performs internal refinement of features for each flow type:
\begin{align}
\mathbf{Q}_i &= \mathbf{h} \mathbf{W}_Q, \quad
\mathbf{K}_i = \mathbf{h} \mathbf{W}_K, \quad
\mathbf{V}_i = \mathbf{h} \mathbf{W}_V,
\end{align}
where $\mathbf{W}_Q \in \mathbb{R}^{d_h \times d_q}$, $\mathbf{W}_K \in \mathbb{R}^{d_h \times d_k}$, $\mathbf{W}_V \in \mathbb{R}^{d_h \times d_v}$ are learnable weight matrices, and $d_q$, $d_k$, $d_v$ are the dimensions of query, key, and value, respectively. The self-attention weights are computed through the softmax function:
\begin{align}
\mathbf{W}^{self-attention}_i = \mathit{softmax}\left(\frac{\mathbf{Q}_i \mathbf{K}_i^\top}{\sqrt{d_k}}\right),
\end{align}
where $d_k$ is the key dimension, used to scale attention scores and prevent gradient vanishing. The final self-attention output is:
\begin{align}
\mathbf{h}_i^{self} = \mathbf{W}^{self-attention}_i \mathbf{V}_i.
\end{align}
The self-attention module can capture long-range dependencies within flow types and identify important physical features.

\vspace{1pt}
\textbf{\ding{173} Cross-Attention Module.} The cross attention module achieves knowledge interaction between flow types. For flow type $f_i$, cross-attention is computed between its query vector and key/value projections associated with the other flow types as follows:
\begin{equation}
\mathbf{Q}_i^{cross} = \mathbf{h}_i^{self} \mathbf{W}_Q^{(i)},\quad
\mathbf{K}_j^{cross} = \mathbf{h}_i^{self} \mathbf{W}_K^{(j)},\quad
\mathbf{V}_j^{cross} = \mathbf{h}_i^{self} \mathbf{W}_V^{(j)},
\end{equation}
where $j \in \{1, 2, \ldots, N\}$ represents the index of other flow types (corresponding to flow type $f_j$), $N$ is the total number of flow types, and $\mathbf{W}_Q^{(i)} \in \mathbb{R}^{d_h \times d_{q}^{cross}}$, $\mathbf{W}_K^{(j)} \in \mathbb{R}^{d_h \times d_{k}^{cross}}$, $\mathbf{W}_V^{(j)} \in \mathbb{R}^{d_h \times d_{v}^{cross}}$ are learnable weight matrices for cross attention, with $d_{q}^{cross}$, $d_{k}^{cross}$, and $d_{v}^{cross}$ being the dimensions of cross attention query, key, and value, respectively. The projection matrices $\mathbf{W}_K^{(j)}$ and $\mathbf{W}_V^{(j)}$ are specific to each flow type, parameterized by flow type $f_j$ and learned from flow type $f_j$'s data during joint training, thereby encoding inductive biases specific to flow type $f_j$. The cross-attention weights are computed as:
\begin{align}
\mathbf{W}^{cross-attention}_{i,j} = \mathit{softmax}\left(\frac{\mathbf{Q}_i^{cross} (\mathbf{K}_j^{cross})^\top}{\sqrt{d^{cross}_k}}\right),
\end{align}
where $d^{cross}_k$ is the key dimension for cross attention. The final cross-attention output is the weighted sum of all flow-type features:
\begin{align}
\mathbf{h}_i^{cross} = \sum_{j\neq i} \mathbf{W}^{cross-attention}_{i,j} \mathbf{V}_j^{cross}.
\end{align}

\vspace{1pt}
\textbf{\ding{174} Feature Fusion.} To balance self-features and cross-flow information, a weighted fusion strategy is adopted:
\begin{align}
\mathbf{h}_i^{enhanced} = \alpha \mathbf{h}_i^{self} + (1-\alpha) \mathbf{h}_i^{cross},
\end{align}
where $\alpha \in [0,1]$ is a learnable parameter controlling the relative importance of the two types of information. This design enables the network to dynamically adjust its dependence on self-features and cross-flow information during training. The enhanced feature $\mathbf{h}_i^{enhanced}$ enables the network to selectively aggregate fluid primitives relevant to the current flow type from the shared feature pool, thereby mitigating interference from irrelevant patterns.

\vspace{6pt}
\noindent\textbf{2) Task Specific Dedicated Layers} 
\vspace{1pt}

The multi-task attention fusion features $\mathbf{h}_i^{enhanced}$ are then fed into the task-specific dedicated layers. Each flow type $f_i$ has its own independent dedicated head $\mathcal{N}_i$, which outputs the stream function and pressure:
\begin{align}
(\psi_i, p_i) = \mathcal{N}_i(\mathbf{h}_i^{enhanced}; \Theta_i).
\end{align}
where $\psi_i$ and $p_i$ are the stream function and pressure field of flow $i$, respectively. $\Theta_i$ represents the task-specific parameters for flow type $i$, the velocity components are obtained from the stream function via automatic differentiation,
\begin{align}
\mu_i = \frac{\partial \psi_i}{\partial y},\quad \nu_i = -\frac{\partial \psi_i}{\partial x},
\end{align}
where $\mu_i$ and $\nu_i$ are respectively the velocity components of the stream function $\psi_i$ in the $x$ and $y$ directions, and $p_i$ is directly predicted by the dedicated head. In this way, we report the final flow-specific predictions as $(\mu_i, \nu_i, p_i)$,

\subsubsection{Dynamic Weight Allocation and Optimization}

A key challenge in multi-flow learning is how to balance loss weights across different flow types. Traditional fixed-weight methods cannot adapt to the convergence characteristics of different flow types. When training heterogeneous fluid topologies in parallel, differences in underlying physical mechanisms lead to imbalanced optimization landscapes, where large-gradient tasks may dominate the parameter update direction, causing slow convergence on complex tasks, known as ``gradient pathology''. Therefore, we propose an adaptive weight-balancing strategy based on dynamic weight adjustment.

\vspace{6pt}
\noindent\textbf{1) Loss Function Components} 
\vspace{1pt}

For each flow type $f_i$, the total loss consists of three main components:
\begin{align}
\mathcal{L}_i = \mathcal{L}_{eq,i} + \mathcal{L}_{bc,i} + \mathcal{L}_{data,i},
\end{align}
where $\mathcal{L}_{eq,i}$ is the equation residual loss, $\mathcal{L}_{bc,i}$ is the boundary condition loss, and $\mathcal{L}_{data,i}$ is the data supervision loss.

\vspace{1pt}
\textbf{\ding{172} Equation Residual Loss.} The equation residual loss is defined as:
\begin{align}
\mathcal{L}_{eq,i} = \frac{1}{|\Omega_i|} \int_{\Omega_i} \left|\frac{\partial \mathbf{u}_i}{\partial t} + \mathbf{u}_i \cdot \nabla \mathbf{u}_i + \frac{1}{\rho_i}\nabla p_i - v_i \nabla^2 \mathbf{u}_i - \bm{\delta}_i\right|^2 dx,
\end{align}
where $|\Omega_i|$ is the area of domain $\Omega_i$.

\vspace{1pt}
\textbf{\ding{173} Boundary Condition Loss.} The boundary condition loss is defined as:
\begin{align}
\mathcal{L}_{bc,i} = \frac{1}{|\partial \Omega_i|} \int_{\partial \Omega_i} \left[|\mathbf{u}_i - \mathbf{u}_{i,bc}|^2 + |p_i - p_{i,bc}|^2\right] ds,
\end{align}
where $\partial \Omega_i$ is the boundary of domain $\Omega_i$, $|\partial \Omega_i|$ is the length of the boundary, and $\mathbf{u}_{i,bc}$ and $p_{i,bc}$ are the given values on the boundary.

\vspace{1pt}
\textbf{\ding{174} Data Supervision Loss.} The data supervision loss is defined as:
\begin{align}
\mathcal{L}_{data,i} = \frac{1}{|D_i|} \sum_{(x,t) \in D_i} \left[|\mathbf{u}_i(x,t) - \mathbf{u}_{i,obs}(x,t)|^2 + |p_i(x,t) - p_{i,obs}(x,t)|^2\right],
\end{align}
where $D_i$ is the observation dataset for flow type $f_i$, $|D_i|$ is the number of observation points, and $\mathbf{u}_{i,obs}$ and $p_{i,obs}$ are the observed values.

\vspace{6pt}
\noindent\textbf{2) Dynamic Weight Calculation} 
\vspace{1pt}

To mitigate the ``gradient pathology'' issue, this paper proposes a Dynamic Weight Allocation (DWA) strategy based on task learning progress to calculate weights $\lambda_i(t)$ in real-time. DWA measures the training difficulty of each flow type based on its recent loss decline rate and adjusts its weight accordingly to balance the effective learning rates across tasks. First, the relative improvement rate is defined as:
%The core idea of adaptive weights is to dynamically adjust weights based on the training progress of each flow type. The relative improvement rate is defined as:
\begin{align}
r_i^{(t)} = \frac{\mathcal{L}_i^{(t-1)} - \mathcal{L}_i^{(t)}}{\mathcal{L}_i^{(t-1)}},
\end{align}
which reflects the training progress of flow type $f_i$ at time step $t$. A larger $r_i(t)$ value indicates a slower recent decline in loss for that task. Based on the relative improvement rate, the adaptive weights are computed as:
\begin{align}
\lambda_i^{(t)} = \frac{\exp(r_i^{(t)} / \tau)}{\sum_{j=1}^i \exp(r_j^{(t)} / \tau)},
\end{align}
where $\tau > 0$ is the temperature parameter controlling the smoothness of the weight distribution: when $\tau \to \infty$, the distribution tends to be uniform; when $\tau \to 0$, weights tend to concentrate on the task with the slowest current loss decline. By selecting an appropriate $\tau$, DWA achieves a trade-off between focusing on difficult tasks and maintaining overall training stability. In the initial phase of training ($t \le 2$), we initialize all weights as $\lambda_i^{(t)}=1$.

%sharpness of the weight distribution, and $j = 1, 2, \ldots, K$ represents the index of all flow types. When $\tau$ is large, the weight distribution tends to be uniform; when $\tau$ is small, weights are more concentrated on flow types with faster training progress.

\textbf{Weight Smoothing Mechanism.} To avoid violent weight fluctuations, an exponential moving average mechanism is introduced:
\begin{align}
\tilde{\lambda}_i^{(t)} = \gamma \lambda_i^{(t-1)} + (1-\gamma) \lambda_i^{(t)},
\end{align}
where $\gamma \in [0,1]$ is the smoothing parameter, typically set to 0.9. This mechanism ensures smooth weight changes and improves training stability.

\textbf{Multi-Flow Total Loss.} The final multi-flow total loss is computed as:
\begin{align}
\mathcal{L}_{total} = \sum_{i=1}^N \tilde{\lambda}_i^{(t)} \mathcal{L}_i.
\end{align}

\subsubsection{Training Process and Algorithm}
The UniPINN framework is trained end-to-end using a unified multi-task optimization loop that jointly minimizes the physics-informed losses across all flow types. The complete training procedure is summarized in Algorithm~\ref{alg:unipinns_training}. During each iteration, the shared backbone extracts common physical features from input coordinates, which are then refined by the cross-flow attention mechanism and decoded through flow-specific dedicated layers to produce predictions for each flow type. The total loss is computed as a weighted sum of the Navier-Stokes residuals, boundary condition residuals, and initial condition residuals across all flows, with adaptive weights updated based on gradient statistics or loss magnitudes.

Throughout training, several key indicators are monitored in real time to ensure stability and effectiveness, including the loss trajectories of individual flows, the evolution of adaptive weights, and the attention patterns across flow types. If signs of training instability are detected, such as diverging losses, vanishing gradients, or oscillatory weight assignments, the learning rate or weight balancing hyperparameters are adjusted accordingly. This monitoring mechanism safeguards against negative transfer and maintains balanced learning across heterogeneous flow regimes.

%The training process ensures that the network can simultaneously learn common physical features across flow types while maintaining flow-specific characteristics. The adaptive weight balancing strategy is used to optimize the complete UniPINN framework within a unified multi-task training loop. During training, the convergence of each flow type is monitored, and weight allocation is dynamically adjusted.

%The algorithmic steps are summarized in Algorithm~\ref{alg:unipinns_training} below. Throughout the training process, the loss changes, weight distributions, and attention patterns of each flow type are monitored in real-time to ensure the stability and effectiveness of the training process. When training instability is detected, the learning rate or weight balancing parameters are adjusted.

\begin{algorithm}[H]
\caption{UniPINN Multi-Task Training Processing}
\label{alg:unipinns_training}
\begin{algorithmic}[1]
\Require Flow type set $\mathcal{F} = \{f_1, \ldots, f_N\}$; max epochs $E_{\max}$; hyperparameters $\eta$, $\tau$, $\gamma$.
\Ensure Optimized parameters $\Theta^* = \{\Theta_{shared}, \{\Theta_i\}_{i=1}^N\}$.

\State Initialize shared parameters $\Theta_{shared}$ and task-specific parameters $\{\Theta_i\}_{i=1}^N$.
\State Initialize smoothed DWA weights $\tilde{\lambda}_i^{(0)} \leftarrow 1$ for all $i=1,\ldots,N$.

\For{epoch $e = 1, \ldots, E_{\max}$}
    \State Sample mini-batches from $\Omega_i$ and $\partial\Omega_i$ for all $f_i \in \mathcal{F}$.
    \State Set aggregated batch loss $\mathcal{J}_{batch} \leftarrow 0$.

    \For{each flow type $f_i \in \mathcal{F}$}
        \State Computer $(\mu_i, \nu_i, p_i)$ via shared backbone $\mathcal{N}_s$ and attention module $\mathcal{N}_i$.
        \State Compute total loss $\mathcal{L}_{total}$ via automatic differentiation.
        \State Accumulate $\mathcal{J}_{batch} \mathrel{+}= \tilde{\lambda}_i^{(e-1)} \, \mathcal{L}_i$.
    \EndFor

    \State Compute gradients $\nabla_{\Theta} \mathcal{J}_{batch}$ and update $\Theta$ using Adam with learning rate $\eta$.

    \If{$e \ge 2$}
        \State Compute relative improvement $r_i^{(e)}$ from recent $\{\mathcal{L}_i\}$ history.
        \State Compute raw weights $\lambda_i^{(e)}$ using DWA softmax with temperature $\tau$.
        \State Update smoothed weights $\tilde{\lambda}_i^{(e)} = \gamma \lambda_i^{(e-1)} + (1-\gamma) w_i^{(e)}$.
    \EndIf
\EndFor

\State \Return $\Theta^*$.
\end{algorithmic}
\end{algorithm}

\section{Experiments}
\label{sec:experiments_revised}
In this section, we evaluate UniPINN on three canonical flow types: lid-driven cavity flow, Poiseuille pipe flow, and Couette flow, under varying Reynolds numbers, and compare against several baselines using MSE. Qualitative visualizations illustrate flow field fidelity. Ablation studies analyze the contribution of each component, while convergence and transferability analyses further validate the framework's effectiveness.

\subsection{Datasets and Preprocessing}

\vspace{6pt}
\noindent\textbf{1) Dataset Description} 
\vspace{1pt}

We consider three distinct laminar flow types from the set $\mathcal{F} = \{f_1, f_2, f_3\}$, corresponding to lid-driven cavity flow, Poiseuille pipe flow, and Couette flow. Owing to their fundamental differences in driving mechanisms, geometric domains, and boundary conditions, these tasks present significant challenges for unified modeling:

\begin{itemize}
\item 
\textbf{Lid-driven cavity flow} is simulated within a unit square domain $\Omega = [0, 1]^2$. The flow is driven by the top wall moving at a constant horizontal velocity $u_{top} = 1$, while the other three stationary walls impose no-slip conditions, i.e., $\mathbf{u} = \mathbf{0}$. The primary challenge lies in capturing the strong shear layer at the top alongside the formation of the primary central vortex and secondary corner vortices.

\item 
\textbf{Poiseuille pipe flow} is simulated in a rectangular channel. The flow is driven by a pressure gradient, characterized by a parabolic velocity profile imposed at the inlet and an open outflow condition at the outlet. The top and bottom walls enforce no-slip conditions. This task tests the model's ability to handle open boundary systems and develop fully developed flow profiles.

\item 
\textbf{Couette flow} occurs between two parallel infinite plates. The flow is driven purely by viscous drag resulting from the moving top plate velocity $U$ relative to the stationary bottom plate. This shear-dominated flow exhibits a linear velocity profile in steady state and serves as a basic benchmark for testing viscous stress resolution.
\end{itemize}

For all flow types, high-fidelity numerical solutions from PDEBench~\citep{takamoto2022pdebench} serve as the ground truth for evaluation. The vast differences in boundary topology (e.g., closed vs.\ open) and dominant physical forces (e.g., shear-driven vs.\ pressure-driven) among these tasks create an ideal testbed for evaluating the robustness of our multi-task learning framework.

\vspace{6pt}
\noindent\textbf{2) Data Preprocessing} 
\vspace{1pt}

Different flow types exhibit different geometric configurations and physical parameters, leading to significant differences in the scale and distribution of input data. To ensure that the shared backbone can effectively handle these differences, we adopt the following input normalization and coordinate transformation strategies.

\textbf{Coordinate Normalization.} For each flow  $f_i$, the spatial domain is $\Omega_i = [x_{min,i}, x_{max,i}]$ $\times$ $[y_{min,i}, y_{max,i}]$. Coordinates are normalized to the standard interval $[0,1]^2$:
\begin{align}
\tilde{x} &= \frac{x - x_{min,i}}{x_{max,i} - x_{min,i}}, \quad
\tilde{y} = \frac{y - y_{min,i}}{y_{max,i} - y_{min,i}},
\end{align}
where $x_{min,i}$, $x_{max,i}$, $y_{min,i}$, and $y_{max,i}$ are the minimum and maximum coordinate values of flow type $f_i$ in the $x$ and $y$ directions, respectively. This normalization eliminates geometric scale differences between flow types, enabling the network to focus on learning physical patterns rather than geometric scales. 

\textbf{Time Normalization.} Time normalization is performed based on the characteristic time scale $T_{char,i} = L_i / U_i$ of each flow type $f_i$:
\begin{align}
\tilde{t} = \frac{t}{T_{char,i}},
\end{align}
where $L_i$ and $U_i$ are the characteristic length and velocity scales. This ensures that the time evolution of different flow types occurs on a comparable scale.

\textbf{Physical Parameter Encoding.} To fully utilize flow-specific physical parameters, they are encoded as additional input features:
%Moreover, flow-specific physical parameters are encoded as additional inputs:
\begin{align}
\mathbf{z}_i = [v_i, \rho_i, Re_i],
\end{align}
where $Re_i = U_i L_i / v_i$ is the Reynolds number. 

\textbf{Feature Enhancement.} All the features are concatenated with the normalized coordinates and fed to the shared backbone. To further improve the network's expressive capability, periodic feature enhancement is applied:
\begin{align}
\mathbf{x}^{enhanced}_{in} = [\tilde{x}, \tilde{y}, \tilde{t}, \sin(2\pi\tilde{x}), \cos(2\pi\tilde{x}), \sin(2\pi\tilde{y}), \cos(2\pi\tilde{y}), \mathbf{z}_i],
\end{align}
which helps the network capture periodic structures in the flow field.

\subsection{Experimental Setup and Baselines}

We describe the architecture, training protocol, evaluation metrics, and baseline methods used to validate the proposed framework.

The shared backbone $\mathcal{N}_s$ and each task-specific dedicated head $\mathcal{N}_i$ are implemented as fully connected MLPs with multiple hidden layers. Consistent with the methodology, the hyperbolic tangent ($\tanh$) activation function is used so that the high-order derivatives required for PDE residuals are well-defined. The task-specific feature extraction stage employs a multi-head cross-flow attention module with multiple parallel heads. Training is performed with the Adam optimizer and a standard learning rate decay schedule; the dynamic weight allocation (DWA) strategy is applied to balance the multi-flow loss during training. The mean squared error (MSE) over all spatial collocation points is adopted as the primary quantitative metric to evaluate the prediction accuracy of the velocity field $\mathbf{u}_i$ and the pressure field $p_i$ for each flow type $f_i$.

To establish a comprehensive evaluation benchmark, UniPINN is compared with baselines at three levels.

\begin{itemize}
\item Foundational data-driven baselines include linear regression, Gaussian process regression, and pure deep neural networks (DNN); these methods rely only on data and provide a reference for prediction without physical constraints.

\item Standard PINN for single-flow task, each trained on that flow type alone; they offer an isolated-task performance reference and allow assessment of whether multi-task training leads to negative transfer.

\item Advanced PINN variants and deep solvers include LAAF-PINN (Locally Adaptive Activation Functions), which uses learnable local activations to improve expressiveness; KIH-PINN (Knowledge-Integrated), which incorporates prior physical knowledge into the network; ALPINN (Attention-based PINN), which employs attention mechanisms for physics-informed learning; and MMPDE-Net (Multi-Modal PDE Net), which relies on parallel solution strategies and multi-scale architectures for PDE systems. Comparisons with these architectures highlight the advantages of UniPINN in handling highly heterogeneous physical tasks.
\end{itemize}

\subsection{Comparative Results}
\subsubsection{Quantitative Accuracy Analysis}
To ensure a fair comparison, the single-task PINN baselines are configured to match the depth and width of the corresponding task branch in UniPINN. These single-task models are trained on a combined dataset containing samples from all three flow types, and then evaluated separately on each flow. This setup ensures that any performance difference can be attributed to the multi-task architecture rather than variations in network capacity or training data. Notably, because UniPINN shares a common backbone across flow types, it achieves substantial parameter efficiency; training three independent single-task PINNs requires considerably more parameters than the proposed unified model. Table~\ref{tab:main_comparison} reports the mean squared error (MSE) of the predicted velocity and pressure fields evaluated over spatial collocation points for all three flow regimes.

%To ensure a fair comparison for the effectiveness of the multi-task framework, the single-task PINN method was trained on a mixed dataset of the three types of flow, and then tested on each of them separately. Moreover, the single-task PINN baselines are configured to match the depth and width of a single task branch within UniPINN. UniPINN achieves substantial parameter efficiency by sharing a backbone across flow types rather than maintaining separate networks per task; training three separate single-task PINNs, therefore, requires more parameters than UniPINN. Table~\ref{tab:main_comparison} reports the mean squared error (MSE) of predicted velocity and pressure over spatial collocation points for all three flow types.

\begin{table}[h]
\centering
\caption{Comparative Study Results: Performance Comparison (MSE) Across Different Methods.}
\label{tab:main_comparison}
\renewcommand{\arraystretch}{1}
\setlength{\tabcolsep}{20pt} 
\resizebox{1\textwidth}{!}{
\begin{tabular}{lccc}
\toprule
\textbf{Method} & \textbf{Lid-Driven} & \textbf{Pipe Flow} & \textbf{Couette flow} \\
\midrule
Linear Regression\citep{das2025cfd} & 6.61e-02 & 5.79e-01 & 6.61e-02 \\
GPR\citep{yang2020uncertainties} & 2.51e-02 & 3.97e-01 & 4.95e-01 \\
DNN\citep{wang2024reduced} & 3.25e-02 & 2.80e-01 & 6.83e-02 \\
\midrule
LAAF-PINN\citep{XU2025123779} & - & 2.26e-01 & - \\
KIH-PINN\citep{du2024knowledge} & 2.49e-02 & 2.41e-01 & 3.39e-02 \\
AL-PINN\citep{SON2023126424} & 3.12e-02 & 1.91e-01 & 4.73e-02 \\
MMPDE-Net\citep{lopez2022parallel} & 1.68e-02 & 4.26e-01 & 3.78e-02 \\
\midrule
Standard PINN\citep{ref1_raissi2019physics} & 2.87e-02 & 3.36e-01 & 4.68e-02 \\
\midrule
\textbf{UniPINN (Proposed)} & \textbf{1.27e-02} & \textbf{1.25e-01} & \textbf{1.27e-02} \\
\bottomrule
\end{tabular}
}
\end{table}

The comparative results in Table~\ref{tab:main_comparison} demonstrate the clear advantages of the proposed UniPINN framework. Among all evaluated methods, traditional numerical fitting approaches (Linear Regression, GPR, and DNN) consistently yield the highest prediction errors across all three flow types. These data-driven models, lacking any physical constraints, struggle to capture the underlying fluid dynamics. In contrast, the standard PINN achieves improved accuracy over purely data-driven baselines, confirming the benefit of embedding Navier–Stokes residuals into the loss function. However, its performance remains suboptimal, with errors of 2.87e-02, 3.36e-01, and 4.68e-02 on lid-driven flow, pipe flow, and Couette flows, respectively.

Recent PINN variants, including LAAF-PINN, KIH-PINN, AL-PINN, and MMPDE-Net, further enhance predictive accuracy through architectural innovations or adaptive training strategies. Nevertheless, a closer examination reveals that these single-task methods exhibit uneven performance across different flow regimes. For instance, MMPDE-Net ranks second on the lid-driven flow (1.68e-02), but its error on pipe flow is higher (4.26e-01), even exceeding that of the standard PINN. Similarly, AL-PINN achieves the second-lowest error on pipe flow (1.91e-01), but its performance on Couette flow (4.73e-02) is only middling. This inconsistency reflects a fundamental limitation of single-task approaches: each is tailored to a specific flow regime and fails to generalize robustly across diverse physical scenarios.

In contrast, UniPINN attains the lowest MSE on all three flow types, with errors of 1.27e-02, 1.25e-01, and 1.27e-02, respectively. Compared to the next-best method for each flow, UniPINN reduces the error by 24.4\% on lid-driven flow, 34.6\% on pipe flow, and 62.5\% on Couette flow. These substantial improvements underscore the effectiveness of the proposed shared-specialized architecture and cross-flow attention mechanism, which enable the model to leverage common physical features across flows while preserving regime-specific characteristics. By unifying multi-flow learning, UniPINN not only outperforms all baselines but also delivers balanced and consistent accuracy, validating its potential for generalizable physics-informed modeling.

\subsubsection{Qualitative Visualization Analysis}

\begin{figure}[H]
    \centering
    \includegraphics[width=\linewidth]{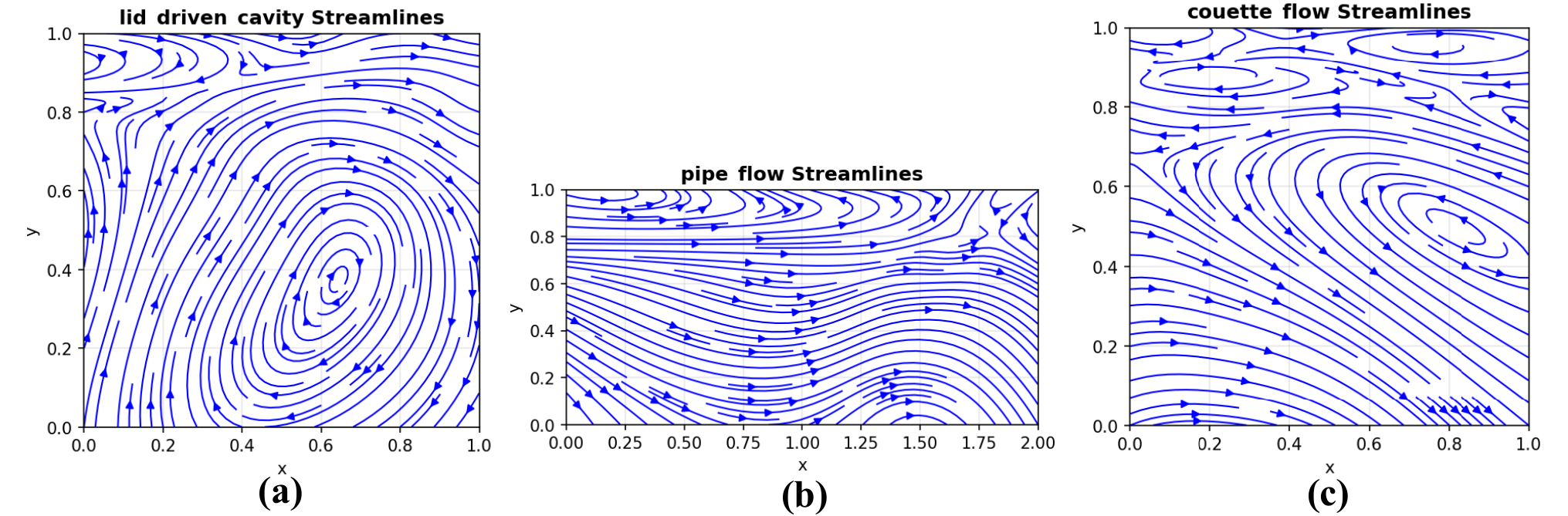}
    \vspace{-24pt}
    \caption{Predicted flow field visualization (streamline plot). The model accurately captures characteristic physical structures: (a) Lid-driven cavity primary vortex; (b) Pipe flow parabolic profile; (c) Couette flow linear profile.}
    \label{fig:streamlines_revised}
\end{figure}

Figure~\ref{fig:streamlines_revised} visualizes the predicted flow fields via streamline plots for the three flow types. As shown in the figure, UniPINN reproduces the characteristic structures of each regime: in lid-driven cavity flow, the primary central vortex and corner recirculation zones are well captured, consistent with the top-wall driving and no-slip boundaries; in pipe flow, the developing parabolic velocity profile and streamwise alignment reflect the pressure-driven inlet-outlet configuration; in Couette flow, the linear shear profile between the moving and stationary walls is recovered. These patterns differ sharply in geometry, boundary conditions, and driving mechanisms, so their accurate reproduction by a single unified model indicates that the shared backbone and task-specific heads successfully disentangle flow-type-specific features without cross-task interference. The smoothness and physical consistency of the streamlines (no spurious crossings or discontinuities) further suggest that the predictions satisfy the continuity and momentum constraints implied by the Navier-Stokes formulation, rather than merely fitting data in a non-physical way.

\subsection{Ablation Study}
\label{sec:ablation}
\subsubsection{Component Analysis}
Table~\ref{tab:ablation} summarizes the effect of removing each core component (cross-flow attention, input normalization, and DWA) from UniPINN. The full model achieves the lowest MSE on all three flow types; removing any one component leads to a clear performance drop, and removing multiple components degrades results further.

\begin{table}[htbp]
\centering
\caption{Ablation study results: Component analysis and performance comparison (MSE).}
\label{tab:ablation}
\renewcommand{\arraystretch}{1}
\setlength{\tabcolsep}{10pt} 
\resizebox{1\textwidth}{!}{
\begin{tabular}{cccccc}
\toprule
\textbf{Attention} & \textbf{Input Norm.} & \textbf{DWA} & \textbf{Lid-Driven} & \textbf{Pipe Flow} & \textbf{Couette flow} \\
\midrule
\yes & \yes & \yes & \textbf{1.27e-02} & \textbf{1.25e-01} & \textbf{1.27e-02} \\
\yes & \yes & \no & 1.83e-02 & 2.17e-01 & 1.18e-01 \\
\yes & \no & \yes & 2.57e-02 & 1.65e-01 & 2.40e-02 \\
\no & \yes & \yes & 8.59e-02 & 2.62e-01 & 1.62e-02 \\
\no & \no & \yes & 1.22e-01 & 3.51e-01 & 6.39e-02 \\
\no & \no & \no & 1.42e-01 & 3.78e-01 & 9.21e-02 \\
\bottomrule
\end{tabular}
}
\end{table}

Specifically, removing DWA causes a substantial relative increase in pipe flow (about 73.6\% higher MSE than the full model), and across all three flows, the summed MSE grows by roughly 135\%, consistent with the need to balance tasks that have different convergence behavior under open versus closed boundary conditions. Removing the cross-flow attention mechanism leads to the most severe overall degradation: the lid-driven cavity error increases by about 576.4\%, while pipe and Couette flow see relative increases of 109.6\% and 27.6\%, respectively; in total, the summed MSE increases by about 142.1\%, indicating that the attention module is critical for mitigating negative transfer and allowing each task to focus on relevant features from the shared backbone. Omitting input normalization roughly doubles errors on lid-driven cavity and Couette flow (increases of 102.4\% and 89.0\%, respectively) and raises pipe-flow MSE by 32.0\%; overall, the summed MSE increases by 42.8\%. These results support that all three components contribute in complementary ways to the final performance and that removing any of them consistently harms at least one flow type by 30\%–580\%.

\begin{figure}[t!]
    \centering
    \includegraphics[width=\linewidth]{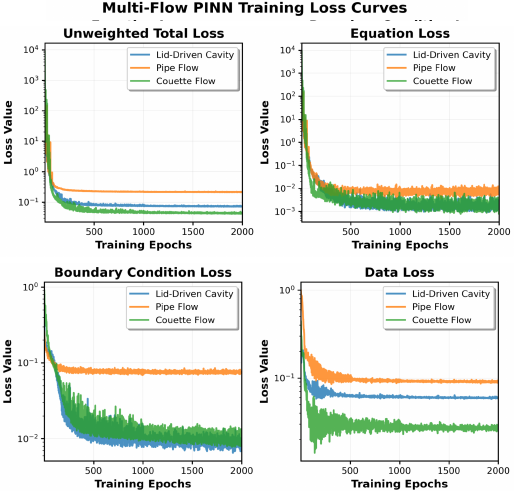}
    \vspace{-24pt}
    \caption{Training convergence analysis (loss curves) during multi-flow PINN training. (a) Unweighted Total Loss; (b) Equation Loss (Navier-Stokes residuals); (c) Boundary Condition Loss; (d) Data Loss. All loss components and the total loss show stable convergence under the DWA strategy.}
    \label{fig:loss_curves_revised}
\end{figure}

\subsubsection{Convergence Analysis}

Figure~\ref{fig:loss_curves_revised} shows the evolution of the main loss terms during multi-flow training. From the figure we can see that the unweighted total loss, equation loss (Navier-Stokes residuals), boundary condition loss, and data loss all decrease rapidly in the early phase and then stabilize for all three flow types. Pipe flow generally exhibits higher loss levels than lid-driven cavity and Couette flow, consistent with its more complex Neumann boundary conditions and inlet-outlet configuration; Couette flow converges to the lowest levels, in line with its simpler linear shear profile. No sustained oscillation or divergence is observed, indicating that the dynamic weight allocation (DWA) strategy effectively balances the contributions of the three tasks and avoids gradient dominance by any single flow type. The curves thus support that UniPINN achieves stable and coordinated convergence across heterogeneous flow regimes under joint training.

\begin{table}[htbp]
\centering
%\small
\caption{Cross-task transfer learning performance using pre-trained shared backbone.}
\label{tab:transfer_experiments}
\renewcommand{\arraystretch}{1}
\setlength{\tabcolsep}{5pt} 
\resizebox{1\textwidth}{!}{
\begin{tabular}{cccccc}
%\begin{tabularx}{\textwidth}{>{\raggedright\arraybackslash}X >{\raggedright\arraybackslash}p{2.5cm} >{\centering\arraybackslash}p{1.7cm} >{\centering\arraybackslash}p{1.8cm} >{\centering\arraybackslash}p{1.5cm}}
\toprule
\textbf{Transfer Scenario} & \textbf{Setting} & \textbf{Final MSE} & \textbf{Loss Reduction} & \textbf{Initial Loss} \\
\midrule
\multirow{3}{*}{\makecell{Source: Pipe \& Lid\\Target: Couette}}
  & From scratch   & 0.1226 & Baseline & 3004    \\
  & Frozen Backbone & 0.0236 & 80.7\%    & 1.51    \\
  & Fine-tuned     & \textbf{0.0216} & \textbf{82.4\%}    & 1.08    \\
\midrule
\multirow{3}{*}{\makecell{Source: Pipe \& Couette\\Target: Lid-Driven}}
  & From scratch   & 6.2407 & Baseline & 97.7    \\
  & Frozen Backbone & 0.0268 & 99.96\%   & 4.84    \\
  & Fine-tuned     & \textbf{0.0214} & \textbf{99.97\%}   & 1.76    \\
\midrule
\multirow{3}{*}{\makecell{Source: Couette \& Lid\\Target: Pipe}}
  & From scratch   & 0.2459 & Baseline & 2985.8 \\
  & Frozen Backbone & 0.1410 & 42.6\%    & 2.94    \\
  & Fine-tuned     & \textbf{0.1363} & \textbf{44.6\%}    & 0.95    \\
\bottomrule
\end{tabular}
}
\end{table}

\subsubsection{Cross-Task Transfer Ability Analysis}

A core advantage of our unified framework is the transfer of learned physical knowledge to new flow types. To verify this advantage, we also conduct two experiments to assess cross-task transfer ability.

First, we pre-train the shared backbone on two source tasks and then use it to initialize the UniPINN for the third task, under either frozen-backbone (only heads trained) or fine-tuned (all parameters trainable) settings; Table~\ref{tab:transfer_experiments} compares these to training the target task from scratch. Across all three transfer directions, using the pre-trained backbone reduces the initial loss of the target task by roughly 98\%–99.97\% and yields lower final MSE in every scenario. For transfer to lid-driven cavity flow, the fine-tuned model achieves a loss reduction of 99.97\%; for Couette as the target, the reduction is 82.4\%; and for the challenging pipe-flow target, the improvement is 44.6\%. These results indicate that the shared backbone encodes a general Navier-Stokes structure that transfers across flow regimes with different boundary conditions and driving mechanisms.

\subsubsection{Analysis of Negative Transfer Induced by Attention Features} 
The previous experiments verified a key concept of our method: the cross-flow attention mechanism aims to learn highly task-specific features, such as specific boundary layer structures or vortex patterns, rather than general physical representations. Forcing the reuse of these features across mismatched tasks, i.e., when this concept is violated, inevitably leads to significant negative transfer due to differences between the source task feature distributions and the target task's physical needs.

\begin{table}[htbp]
\centering
\caption{Negative transfer analysis when reusing task-specific attention features. Values indicate Final MSE and Relative Increase over Baseline.}
\label{tab:attention_transfer}
\renewcommand{\arraystretch}{1}
\setlength{\tabcolsep}{12pt} 
\resizebox{1\textwidth}{!}{
\begin{tabular}{l c c c c}
\toprule
\textbf{Target Task} & \textbf{Baseline} & \textbf{Source: Lid} & \textbf{Source: Pipe} & \textbf{Source: Couette} \\
\midrule
Lid-Driven & 0.087 & ---            & 0.135 (+55\%) & \textbf{0.158 (+81\%)} \\
Pipe       & 0.235 & 0.297 (+26\%)  & ---           & 0.339 (+44\%) \\
Couette    & 0.049 & 0.056 (+14\%)  & 0.068 (+39\%) & ---            \\
\bottomrule
\end{tabular}
}
\end{table}

Table~\ref{tab:attention_transfer} displays the impact on performance of directly reusing attention features across tasks, confirming this concept. Obvious negative transfer phenomena are observed in all test cases. Specifically, the greater the difference in fluid dynamic mechanisms between source and target tasks, the more severe the performance degradation. For instance, the features of lid-driven cavity flow are primary and corner vortex structures within a closed domain, while Couette flow is a simple linear shear flow between open boundaries. The physical structures of these two flow regimes barely overlap. Therefore, when attempting to use simple shear features from Couette flow to guide the complex lid-driven cavity flow task, MSE surged by 81\% (increasing from 0.087 to 0.158), representing the maximum increase in the table. Conversely, applying complex lid-driven cavity flow features to the simple Couette flow task also led to a 14\% increase in error. Furthermore, although pipe flow (pressure-driven parabolic profile) and Couette flow (shear-driven linear profile) both belong to channel flows, due to different driving mechanisms, mutual use of their attention features similarly resulted in significant error increases, by 44\% and 39\% respectively.

This finding validates the rationality of our architectural design: the backbone network is responsible for learning general physical laws, while the attention mechanism and subsequent layers must remain task-specific to filter out irrelevant physical patterns and focus on topological features unique to each flow.

\section{Discussion}

%\subsection{Contributions}

%Through extensive experiments on three classic two-dimensional incompressible Navier-Stokes benchmarks (lid-driven cavity, pipe flow, and Couette flow), we verify that UniPINN is an effective unified framework for multi-task learning of heterogeneous flows. Across all tasks, UniPINN achieves the lowest MSE, with gains of roughly 8\%–32\% over single-task PINNs and substantial improvements over purely data-driven and advanced PINN baselines. Qualitative visualizations, convergence curves, and transfer-learning experiments further show that the framework enables stable joint training under mixed boundary conditions, mitigates negative transfer, and encodes a reusable Navier-Stokes structure that reduces initial loss and improves accuracy when adapting to new flow types.

\subsection{Limitations and Future Outlook}

Despite its promising performance, UniPINN still has several limitations. First, the current evaluation is restricted to 2D laminar flows; extending the framework to 3D geometries and high-Reynolds-number turbulence will impose much higher computational costs and tighter requirements on multiscale representation, motivating the exploration of more efficient backbones and lightweight attention designs. Second, although PINNs benefit from mesh-free formulations, their accuracy in sharp-gradient regions such as thin boundary layers or shock-like structures can still lag behind carefully tuned classical solvers, suggesting that hybrid schemes where UniPINN acts as an accelerator or corrector for traditional numerical methods are a valuable direction. Finally, this study focuses on incompressible single-phase flows; applying the framework to more complex scenarios, such as compressible, multiphase, or thermo-fluid systems, will be important for assessing its potential as a general-purpose physical solver.

\subsection{Comparison with Neural Operator Methods}
%为什么不用算子，跟算子类方法的比较
In the broader landscape of AI for Science, the evolution of neural operators has catalyzed various complex architectures aimed at transcending the generalization bottlenecks of traditional physics-informed learning. Attention-based operator models, such as OFormer~\citep{hao2023gnot}, and Graph Neural Operators (GNO)~\citep{kovachki2023neural}, attempt to handle irregular geometries by redesigning information aggregation mechanisms~\citep{wu2024transolver}. However, while these architectures enhance geometric flexibility, they typically incur quadratic computational complexity with respect to the number of spatial points, leading to high memory usage and parameter redundancy in high-resolution 3D simulations~\citep{li2023scalable}. Adaptive spectral methods like AFNO~\citep{zhu2025arbitrary} are efficient for large-scale spatio-temporal data, but essentially operate as data-driven black boxes; without explicit physical constraints during training, they can exhibit ``physical hallucinations''~\citep{lian2023physically}, producing solutions that violate conservation laws in data-sparse regions. Classical architectures such as DeepONet~\citep{he2024sequential}, while theoretically appealing as function-space mappers, often struggle with high-gradient phenomena and require very wide networks to maintain accuracy, resulting in severe parameter bloat~\citep{laudato2025neural}. These limitations suggest that current heavyweight neural operators face a difficult trade-off between computational efficiency, physical fidelity, and reliance on dense grid-based data. In contrast, UniPINN adopts a lighter-weight, physics-informed paradigm: by jointly enforcing Navier-Stokes residuals, boundary conditions, and sparse data within a shared-backbone architecture, it maintains physical consistency and achieves robust cross-regime generalization without relying on extremely large models or dense operator sampling.

\section{Conclusion}
\label{sec:conclusion}

This paper proposes UniPINN, a unified physics-informed neural network framework for multi-task learning of diverse Navier-Stokes flows. By combining a shared backbone with task-specific heads, cross-flow attention, input normalization, and dynamic weight allocation, the framework effectively handles heterogeneous geometries and boundary conditions while mitigating negative transfer and training instability. Experiments on three representative incompressible flows show that UniPINN consistently achieves lower MSE than single-task PINNs, data-driven baselines, and advanced PINN variants, and that all core components are necessary according to ablation and transfer studies. These results indicate that UniPINN provides a lightweight yet physically consistent paradigm for learning across multiple flow regimes and offers a promising basis for extension to more complex physical systems.

\section*{Acknowledgment}
This study is funded in part by the National Key Research and Development Program of China (No. 2025YFE0201800), the National Natural Science Foundation of China (No. 62076005), and the Anhui Provincial University Natural Science Research Major Project (No. 2025AHGXZK20027). Anhui Provincial Natural Science Foundation-Outstanding Youth Project, 2408085Y032. The authors acknowledge the High-performance Computing Platform of Anhui University for providing computing resources.

%% Bibliography
%\bibliographystyle{apalike}
\bibliographystyle{elsarticle-num}
\bibliography{cas-refs}

\begin{thebibliography}{10}
\expandafter\ifx\csname url\endcsname\relax
  \def\url#1{\texttt{#1}}\fi
\expandafter\ifx\csname urlprefix\endcsname\relax\def\urlprefix{URL }\fi
\expandafter\ifx\csname href\endcsname\relax
  \def\href#1#2{#2} \def\path#1{#1}\fi

\bibitem{ref1_raissi2019physics}
M.~Raissi, P.~Perdikaris, G.~E. Karniadakis, Physics-informed neural networks:
  A deep learning framework for solving forward and inverse problems involving
  nonlinear partial differential equations, Journal of Computational Physics
  378 (2019) 686--707.

\bibitem{fang2024learning}
Z.~Fang, S.~Wang, P.~Perdikaris, Learning only on boundaries: A
  physics-informed neural operator for solving parametric partial differential
  equations in complex geometries, Neural computation 36~(3) (2024) 475--498.

\bibitem{ref2_bai2020applying}
X.-d. Bai, Y.~Wang, W.~Zhang, Applying physics informed neural network for flow
  data assimilation, Journal of Hydrodynamics 32~(6) (2020) 1050--1058.

\bibitem{ref3_urbanowicz2023navier}
K.~Urbanowicz, A.~Bergant, M.~Stosiak, A.~Deptu{\l}a, M.~Karpenko,
  Navier-stokes solutions for accelerating pipe flow—a review of analytical
  models, Energies 16~(3) (2023) 1407.

\bibitem{ref4_mehta2019discovering}
P.~P. Mehta, G.~Pang, F.~Song, G.~E. Karniadakis, Discovering a universal
  variable-order fractional model for turbulent couette flow using a
  physics-informed neural network, Fractional calculus and applied analysis
  22~(6) (2019) 1675--1688.

\bibitem{jiang2025gradient}
C.~Jiang, N.-Z. Chen, Gradient-free physics-informed neural networks (gf-pinns)
  for vortex shedding prediction in flow past square cylinders, Computers in
  Industry 169 (2025) 104304.

\bibitem{hanrahan2023predicting}
S.~K. Hanrahan, M.~Kozul, R.~D. Sandberg, Predicting transitional and turbulent
  flow around a turbine blade with a physics-informed neural network, in: Turbo
  Expo: Power for Land, Sea, and Air, Vol. 87103, American Society of
  Mechanical Engineers, 2023, p. V13CT32A010.

\bibitem{ren2024physics}
X.~Ren, P.~Hu, H.~Su, F.~Zhang, H.~Yu, Physics-informed neural networks for
  transonic flow around a cylinder with high reynolds number, Physics of Fluids
  36~(3) (2024).

\bibitem{ref5_malek2005mathematical}
J.~M{\'a}lek, K.~R. Rajagopal, Mathematical issues concerning the
  navier--stokes equations and some of its generalizations, in: Handbook of
  differential equations: evolutionary equations, Vol.~2, Elsevier, 2005, pp.
  371--459.

\bibitem{stiasny2023physics}
J.~Stiasny, S.~Chatzivasileiadis, Physics-informed neural networks for
  time-domain simulations: Accuracy, computational cost, and flexibility,
  Electric Power Systems Research 224 (2023) 109748.

\bibitem{ref8tanarro2020effect}
{\'A}.~Tanarro, R.~Vinuesa, P.~Schlatter, Effect of adverse pressure gradients
  on turbulent wing boundary layers, Journal of Fluid Mechanics 883 (2020) A8.

\bibitem{bonfanti2024generalization}
A.~Bonfanti, R.~Santana, M.~Ellero, B.~Gholami, On the generalization of pinns
  outside the training domain and the hyperparameters influencing it, Neural
  Computing and Applications 36~(36) (2024) 22677--22696.

\bibitem{zhu2024online}
J.~Zhu, X.~Lai, X.~Tang, Y.~Zheng, H.~Zhang, H.~Dai, Y.~Huang, Online
  multi-scenario impedance spectra generation for batteries based on
  small-sample learning, Cell Reports Physical Science 5~(8) (2024).

\bibitem{ref15zhang2021survey}
Y.~Zhang, Q.~Yang, A survey on multi-task learning, IEEE transactions on
  knowledge and data engineering 34~(12) (2021) 5586--5609.

\bibitem{ref14chen2024multi}
S.~Chen, Y.~Zhang, Q.~Yang, Multi-task learning in natural language processing:
  An overview, ACM Computing Surveys 56~(12) (2024) 1--32.

\bibitem{li2024physics}
Y.~Li, L.~Liu, Physics-informed neural network-based nonlinear model predictive
  control for automated guided vehicle trajectory tracking, World Electric
  Vehicle Journal 15~(10) (2024) 460.

\bibitem{yu2022gradient}
J.~Yu, L.~Lu, X.~Meng, G.~E. Karniadakis, Gradient-enhanced physics-informed
  neural networks for forward and inverse pde problems, Computer Methods in
  Applied Mechanics and Engineering 393 (2022) 114823.

\bibitem{MAO2026111816}
F.~Mao, J.~Mei, S.~Lu, F.~Liu, L.~Chen, F.~Zhao, Y.~Hu, Pid: Physics-informed
  diffusion model for infrared image generation, Pattern Recognition 169 (2026)
  111816.

\bibitem{YU2026112123}
J.~Yu, Y.~Zhou, R.~Pan, P.~Lai, H.~Yang, Physics–environment interaction
  network for dense crowd behavior recognition, Pattern Recognition 170 (2026)
  112123.

\bibitem{zhao2024comprehensive}
L.~Zhao, et~al., A comprehensive review of physics-informed neural networks for
  fluid mechanics: Advanced architectures and applications, Reviews in Physics
  12 (2024) 100095.

\bibitem{wang2024piratenets}
S.~Wang, P.~Perdikaris, Piratenets: Physics-informed deep learning with
  residual adaptive networks, Journal of Machine Learning Research 25~(24)
  (2024) 1--51.

\bibitem{ref27duibiLAAFxu2025physics}
Y.~Xu, L.~Zhou, Y.~Lu, Y.~Hu, Y.~Zhang, Physics-informed neural networks
  involving unsteady friction for transient pipe flow, Water Research (2025)
  123779.

\bibitem{ref6_hennigh2021nvidia}
O.~Hennigh, S.~Narasimhan, M.~A. Nabian, A.~Subramaniam, K.~Tangsali, Z.~Fang,
  M.~Rietmann, W.~Byeon, S.~Choudhry, Nvidia simnet™: An ai-accelerated
  multi-physics simulation framework, in: International conference on
  computational science, Springer, 2021, pp. 447--461.

\bibitem{moya2024multi}
C.~Moya, G.~Lin, J.~Cohee, Multi-task learning for physics-informed neural
  networks: Application to fluid-structure interaction, Physics of Fluids
  36~(1) (2024) 017105.

\bibitem{wang2021understanding}
S.~Wang, Y.~Teng, P.~Perdikaris, Understanding and mitigating gradient flow
  pathologies in physics-informed neural networks, SIAM Journal on Scientific
  Computing 43~(5) (2021) A3055--A3081.

\bibitem{LI2026108694}
C.~Li, R.~Zeng, Aw-el-pinns: A multi-task learning physics-informed neural
  network for euler-lagrange systems in optimal control problems, Neural
  Networks 199 (2026) 108694.

\bibitem{JI2025111423}
N.~Ji, Y.~Sun, F.~Meng, L.~Pang, Y.~Tian, Variable multi-scale attention fusion
  network and adaptive correcting gradient optimization for multi-task
  learning, Pattern Recognition 162 (2025) 111423.

\bibitem{ref16chen2025multi}
S.~Chen, X.~Zheng, H.~Wu, A multi-rate sensor fusion and multi-task learning
  network for concurrent fault diagnosis of hydraulic systems, Digital Signal
  Processing 156 (2025) 104796.

\bibitem{li2024multitask}
Y.~Li, et~al., Multi-task learning enhanced physics-informed neural network for
  solving fluid-structure interaction equations, in: 2024 IEEE 7th Information
  Technology, Networking, Electronic and Automation Control Conference (ITNEC),
  IEEE, 2024, pp. 1660--1664.

\bibitem{zhang2025self}
Y.~Zhang, et~al., Self-adaptive weight balanced physics-informed neural
  networks for solving complex coupling equations, in: Third International
  Conference on AIGC, Data Computing and IoT (ACDI 2024), Vol. 13555, SPIE,
  2025, p. 135553X.

\bibitem{subramanian2024adaptive}
B.~Subramanian, et~al., Adaptive loss weighting for physics-informed neural
  networks via gradient variance, Journal of Computational Physics 505 (2024)
  112891.

\bibitem{ref21mauricio2023comparing}
J.~Maur{\'\i}cio, I.~Domingues, J.~Bernardino, Comparing vision transformers
  and convolutional neural networks for image classification: A literature
  review, Applied Sciences 13~(9) (2023) 5521.

\bibitem{peng2023linear}
W.~Peng, Z.~Yuan, Z.~Li, J.~Wang, Linear attention coupled fourier neural
  operator for simulation of three-dimensional turbulence, Physics of Fluids
  35~(1) (2023).

\bibitem{aizpurua2025residual}
J.~Aizpurua, et~al., Residual-based attention physics-informed neural networks
  for spatio-temporal ageing assessment, Engineering Applications of Artificial
  Intelligence 139 (2025) 109556.

\bibitem{li2024transformer}
Z.~Li, et~al., Transformer-based physics-informed neural networks for solving
  partial differential equations, Neurocomputing 570 (2024) 127132.

\bibitem{hao2023gnot}
Z.~Hao, Z.~Wang, H.~Su, C.~Ying, Y.~Dong, S.~Liu, Z.~Cheng, J.~Song, J.~Zhu,
  Gnot: A general neural operator transformer for operator learning, in:
  International Conference on Machine Learning, PMLR, 2023, pp. 12556--12569.

\bibitem{ref10zhao2024comprehensive}
C.~Zhao, F.~Zhang, W.~Lou, X.~Wang, J.~Yang, A comprehensive review of advances
  in physics-informed neural networks and their applications in complex fluid
  dynamics, Physics of Fluids 36~(10) (2024).

\bibitem{YANG2026111868}
K.~Yang, G.~Ding, L.~Sun, J.~Chen, G.~Du, Q.~Zheng, T.~Zhang, Mpm-net:
  Multi-task interactive network with progressive multi-granularity learning
  for herbal medicine recognition, Pattern Recognition 169 (2026) 111868.

\bibitem{ref29gu2024physics}
L.~Gu, S.~Qin, L.~Xu, R.~Chen, Physics-informed neural networks with domain
  decomposition for the incompressible navier--stokes equations, Physics of
  Fluids 36~(2) (2024).

\bibitem{takamoto2022pdebench}
M.~Takamoto, T.~Praditia, R.~Leiteritz, D.~MacKinlay, F.~Alesiani,
  D.~Pfl{\"u}ger, M.~Niepert, Pdebench: An extensive benchmark for scientific
  machine learning, Advances in Neural Information Processing Systems 35 (2022)
  1596--1611.

\bibitem{das2025cfd}
A.~Das, S.~T. Sarwar, N.~K. Das, M.~R. Haque, Cfd data-driven machine learning
  approach for enhanced convective heat transfer in nanofluid-filled lid-driven
  chambers, Journal of Thermal Analysis and Calorimetry 150~(17) (2025)
  13701--13731.

\bibitem{yang2020uncertainties}
Y.~Yang, T.~F.~M. Chui, Uncertainties of machine learning in predicting the
  hydrological responses of lid practices, in: Proceedings of the 22nd IAHR-APD
  Congress, 2020, pp. 1--7.

\bibitem{wang2024reduced}
H.~Wang, G.~Jiang, W.~Wang, Y.~Liu, A reduced-order configuration approach for
  the real-time calculation of three-dimensional flow behavior in a pipe
  network, Physics of Fluids 36~(4) (2024).

\bibitem{XU2025123779}
Y.~Xu, L.~Zhou, Y.~Lu, Y.~Hu, Y.~Zhang, Physics-informed neural networks
  involving unsteady friction for transient pipe flow, Water Research 283
  (2025) 123779.

\bibitem{du2024knowledge}
J.~Du, H.~Li, Q.~Liao, J.~Shen, J.~Zheng, Y.~Liang, A knowledge-inspired
  hierarchical physics-informed neural network for pipeline hydraulic transient
  simulation, arXiv preprint arXiv:2409.10911 (2024).

\bibitem{SON2023126424}
H.~Son, S.~W. Cho, H.~J. Hwang, Enhanced physics-informed neural networks with
  augmented lagrangian relaxation method (al-pinns), Neurocomputing 548 (2023)
  126424.

\bibitem{lopez2022parallel}
M.~Lopez, S.~M. Shontz, W.~Huang, A parallel variational mesh quality
  improvement method for tetrahedral meshes based on the mmpde method,
  Computer-Aided Design 148 (2022) 103242.

\bibitem{kovachki2023neural}
N.~Kovachki, Z.~Li, B.~Liu, K.~Azizzadenesheli, K.~Bhattacharya, A.~Stuart,
  A.~Anandkumar, Neural operator: Learning maps between function spaces with
  applications to pdes, Journal of Machine Learning Research 24~(89) (2023)
  1--97.

\bibitem{wu2024transolver}
H.~Wu, H.~Luo, H.~Wang, J.~Wang, M.~Long, Transolver: A fast transformer solver
  for pdes on general geometries, arXiv preprint arXiv:2402.02366 (2024).

\bibitem{li2023scalable}
Z.~Li, D.~Shu, A.~Barati~Farimani, Scalable transformer for pde surrogate
  modeling, Advances in Neural Information Processing Systems 36 (2023)
  28010--28039.

\bibitem{zhu2025arbitrary}
J.~Zhu, W.~Li, H.~Xu, J.~Jiang, Z.~Liu, J.~Zheng, Arbitrary-scale fusion neural
  operator, in: Proceedings of the 33rd ACM International Conference on
  Multimedia, 2025, pp. 1617--1626.

\bibitem{lian2023physically}
Y.~Lian, J.~Shi, C.~N. Jones, Physically consistent multiple-step data-driven
  predictions using physics-based filters, IEEE Control Systems Letters 7
  (2023) 1885--1890.

\bibitem{he2024sequential}
J.~He, S.~Kushwaha, J.~Park, S.~Koric, D.~Abueidda, I.~Jasiuk, Sequential deep
  operator networks (s-deeponet) for predicting full-field solutions under
  time-dependent loads, Engineering Applications of Artificial Intelligence 127
  (2024) 107258.

\bibitem{laudato2025neural}
M.~Laudato, A neural-operator surrogate for platelet deformation across
  capillary numbers, Bioengineering 12~(9) (2025) 958.

\end{thebibliography}

\end{document}